\documentclass{article}
\usepackage[utf8]{inputenc}
\usepackage{graphicx}
\usepackage{amsmath}
\begin{document}

\title{Hoechst Is All You Need: Lymphocyte Classification with Deep Learning}

\author{Jessica Cooper*, In Hwa Um,\\%
Ognjen Arandjelovi\'c and David J Harrison}
\date{%
    {\small *jmc31@st-andrews.ac.uk}\\%
    University of St Andrews\\%
    \today
}

\maketitle

\section{Abstract}

Multiplex immunofluorescence and immunohistochemistry benefit patients by allowing cancer pathologists to identify several proteins expressed on the surface of cells, enabling cell classification, better understanding of the tumour micro-environment, more accurate diagnoses, prognoses, and tailored immunotherapy based on the immune status of individual patients. However, they are expensive and time consuming processes which require complex staining and imaging techniques by expert technicians. Hoechst staining is much cheaper and easier to perform, but is not typically used in this case as it binds to DNA rather than to the proteins targeted by immunofluorescent techniques, and it was not previously thought possible to differentiate cells expressing these proteins based only on DNA morphology. In this work we show otherwise, training a deep convolutional neural network to identify cells expressing three proteins (T lymphocyte markers CD3 and CD8, and the B lymphocyte marker CD20) with greater than 90\% precision and recall, from Hoechst 33342 stained tissue only. Our model learns previously unknown morphological features associated with expression of these proteins which can be used to accurately differentiate lymphocyte subtypes for use in key prognostic metrics such as assessment of immune cell infiltration, and thereby predict and improve patient outcomes without the need for costly multiplex immunofluorescence.

\section{Introduction}
 
Among patients with cancers of the same stage, clinical outcomes vary widely. This is thought to be in large part due to the complex interaction between tumour cells and the immune response of individual patients, as the proportion, location, and sub-type of lymphocytes present in the tissue has been shown to have important implications for patient prognosis~\cite{Bruni2020-as, Mlecnik2018-ww}. There exist proprietary methods to assess immune cell infiltration, which formally quantify CD3+ and CD8+ T cell lymphocytes both in the centre of tumour and in the invasive margin, as proposed by Galon et al.~\cite{Galon2020-um}. Combining their evaluation with T-and-B score (CD8+ T cell and CD20+ B cell) as per Mlecnik et al.\ had significant predictive power for colorectal cancer patient survival~\cite{Van_den_Eynde2018-jy, Mlecnik2018-ww}, and compared to to the latest guidelines of the American Joint Committee on Cancer/ Union for International Cancer Control (AJCC/ UICC) tumour-node-metastasis (TNM) classification, immune cell infiltration evaluation alone has shown superior prognostic value in international studies of stage I-IV colon cancer patients, and has life-saving applications in clinical decision-making~\cite{Tan2020-ic, Taube2020-hm, Raab2000-bq, Parra2017-yr, Bruni2020-as, Galon2020-um, Galon2014-dk}. However, in order to identify the CD3, CD8 and CD20 expressing cells necessary to calculate these valuable metrics, either multiple immunohistochemistry or multiplexed immunofluorescence are required -- both of which are time consuming and expensive protocols~\cite{Van_den_Eynde2018-jy, Angelova2018-wx}. Using contemporary equipment, three simultaneous rounds of immunohistochemistry takes around three hours and costs approximately \$20 in reagents, whilst multiplex immunofluorescence requires ~9 hours and the associated reagents cost upward of \$70 for a single slide. In this work we show that costly protocols of this type are in fact unnecessary -- it is possible to accurately identify CD3, CD8 and CD20 expressing lymphocytes from common and inexpensive stains. Hoechst and DAPI (popular blue fluorescent, nuclear-specific dyes~\cite{Otto1985-kv, Chazotte2011-fc, Chazotte2011-so}) are far cheaper and easier to perform, costing pennies and requiring just ten minutes per slide. DAPI has better photostability, but since our slides could be imaged immediately in this work we use Hoechst 33342 due to its superior signal-to-noise (genuine DNA stain/autofluorescence) ratio.

\subsection{Artificial Intelligence and Digital Pathology}

Deep learning techniques are increasingly used in digital pathology to assist human experts with a range of diagnostic and prognostic tasks. The data used to train these models often takes the form of whole slide images (WSIs) with associated labels. One application of deep learning in this field is \emph{segmentation} -- that is, given some input image, learning to produce class labels for each pixel in that image. For example, given a set of biopsy slides which have been divided into cancerous and non-cancerous regions by a human annotator, we could train a neural network model to label each pixel in the image as cancerous or non-cancerous~\cite{Trebeschi2017-yw, Lal2021-kh, Li2020-lu, Cha2016-hr}. This is distinct from \emph{classification} tasks, in which a single label is provided for an entire slide or region, and pixel-level classification is not required. 

For image segmentation tasks Convolutional Neural Networks (CNNs) are most widely used~\cite{Sultana2020-yp}. One type of CNN is the U-Net~\cite{Ronneberger2015-ii} -- a type of residual neural network~\cite{He-kaim} -- so named for its 'U' shaped architecture. Residual neural networks in general, and the U-Net in particular, are well suited for segmentation tasks as they allow spatial information from the input to propagate directly to the output. Recent work using U-Nets has shown great promise for digital pathology applications~\cite{Oskal2019-rv, Nikolov2018-sd, Kong2020-fi, Mahbod2019-ra, Alom2018-qi, Koga2021-zb}, but publications to date have focused on automating tasks that human experts can already do. In this work we take a different approach, asking instead -- can a deep neural network learn to do something that human experts cannot? 

\section{Materials and Methods}

\begin{figure}[h!]
\centering
\includegraphics[width=\textwidth]{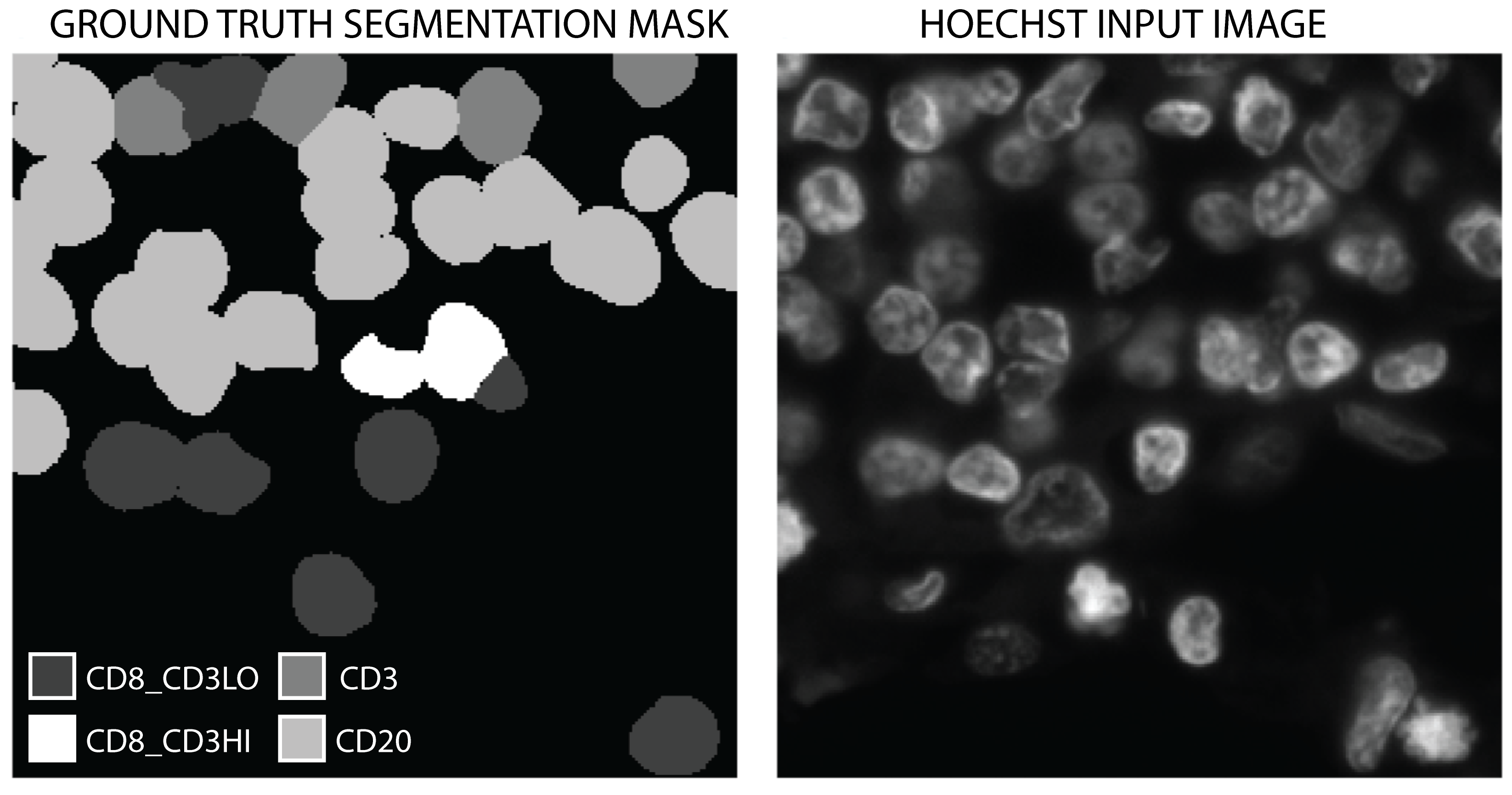}
\caption{Labelled immune cell mask from multiplex immunofluorescence image (left) and corresponding Hoechst 33342 stained patch (right).}
\label{fig:patch_mask_example}
\end{figure}

Our data comprised six Whole Slide Images (WSIs) taken from lung cancer biopsies. These were imaged using Hoechst 33342, and also using multiplex immunofluorescence targeting CD3, CD8 and CD20 expressing immune cells, with a Zeiss Zen Axioscan scanner. We then used an established intensity-based classification technique~\cite{Rizzardi2012-xh} to identify and label cells expressing these proteins in the multiplex immunofluoresence images, the results of which we quality controlled by direct visual inspection to ensure label accuracy. To handle varying levels of co-expression within CD3 and CD8 expressing cells we used a labelling threshold to classify each cell into one of five groups based on protein expression: CD3 only; CD8 and low CD3 (labelled CD8\_CD3LO); CD8 and high CD3 (labelled CD8\_CD3HI); CD20 only; and background/other. We used these labels to create segmentation maps, which we paired with the Hoechst images, as in Figure~\ref{fig:patch_mask_example}.

We used these image pairs to train a semantic segmentation model with U-Net architecture~\cite{Ronneberger2015-ii} and a resnet34 encoder~\cite{He-kaim} to output a segmentation map for each class, taking only a Hoechst image as input.

\subsection{Data Source}

The slides we used were provided by NHS Lothian via David J Harrison. Our six slides were selected for their greater frequency of immune cells from a larger cohort of consenting early-stage lung adenocarcinoma patients.

\subsection{Multiplexed Immunofluorescence (mIF) Protocol}

Leica BOND RX automated immunostainer (Leica Microsystems, Milton Keynes, UK) was utilised to perform mIF. The sections were dewaxed at $72^{\circ}C$ using BOND dewax solution (Leica, AR9222) and rehydrated in absolute alcohol and deionised water, respectively. The sections were treated with BOND epitope retrieval 1 (ER1) buffer (Leica, AR9961) for 20 min at $100^{\circ}C$ to unmask the epitopes. The endogeneous peroxidase was blocked with peroxide block (Leica, DS9800), followed by serum free protein block (Agilent, x090930-2). The sections were incubated with the first primary antibody (CD20, Agilent, M075501-2, 1:400 dilution) for 40 min at room temperature, followed by anti-mouse HRP conjugated secondary antibody (Agilent, K400111-2) for 40 min. Then CD20 antigen was visualised by fluorescein conjugated tyramide signal amplification (TSA) (Akoya Bioscience, NEL741001KT). Redundant antibodies, which were not covalently bound, were stripped off by ER1 buffer at $95^{\circ}C$ for 20 min. Then the second primary antibody (CD8, Agilent, M710301-2, 1:400 dilution) was visualised by TSA Cy3, taking the same steps from peroxide block to the ER1 buffer stripping of the first antibody visualisation. Lastly, the sections were incubated with the last primary antibody (CD3, Agilent, A045229-2, 1:70 dilution), followed by biotinylated anti-rabbit secondary antibodies (Thermo fisher, 65-6140), which was visualised by Alexa flour 750 conjugated streptavidin (Thermo fisher, S21384). Cell nuclei were counterstained by Hoechst 33342 (Thermo fisher, H3570, 1:100) and the sections were mounted with prolong gold antifade mountant (Thermo fisher, P36930).

\subsection{Image Acquisition and Analysis}
Zeiss Axio scan z1 was utilised to capture fluorescent images. Four different fluorescent channels (Hoechst3334, Fluorescein, Cy3 and AF750) were simultaneously used to capture individual channel images under 20x object magnification. The exposure time of four channels (Hoechst33342, Fluorescein, cy3 and AF750) were 8 milliseconds, 20 ms, 50 ms, and 800 ms, respectively. The image was generated in czi format.

The fluorescent images were opened in QuPath v.0.2.3~\cite{Bankhead2017-gv}. StarDist~\cite{Schmidt2018-pg} was utilised to segment cell nuclei using StarDist2D builder. The probability threshold of cell detection, pixel size and the cell expansion was 0.6, 0.2270 and 1.0, respectively. The object classifier was utilised to classify CD20, CD8 and CD3 cells by the intensity threshold of fluorescein, Cy3 and AF750 channels, which were 5000, 4000 and 2200, respectively.

We extracted 7413 $256\times256$ pixel patches with a 50\% overlap from the Hoechst 33342 stained slides at full resolution, and paired them with the per-pixel class labels from the immunofluorescence intensity classifier as shown in figure~\ref{fig:patch_mask_example}. Each Hoechst patch was normalised individually, and the total dataset split into training, validation and testing subsets with ratio 80:10:10, providing 5930 training samples, 741 validation samples, and 742 test samples. We also tested training the model on five of the six slides, retaining the the sixth as a holdout test set in order to ascertain the generalisation ability, and found no decrease in performance.

\begin{table}
    \centering
    \begin{tabular}{|c|c|c|c|c|c|}
        \hline
         & CD8\_CD3LO & CD3 & CD20 & CD8\_CD3HI & Other\\
        \hline
        Count & 14536 & 14480 & 7528 & 1412 & 153811\\
        Presence & 5259 & 4955 & 1352 & 1071 & 7375\\
        Coverage & 7.6\% & 7.5\% & 3.9\% & 0.7\% & 80.2\%\\
        \hline
    \end{tabular}
    \caption{Class representation across the dataset: the total number of cells present (Count); the number of samples containing at least one cell of that type (Presence); and the total pixel percentage of input covered each cell type (Coverage).}
    \label{tab:class_stats}
\end{table}

This dataset exhibited some class imbalance, with most of the total input consisting of background or cells expressing none of the proteins that we are concerned with. CD8\_CD3LO expressing lymphocytes formed the majority of positive examples, closely followed by CD3. CD20 and CD8\_CD3HI were rarer, as shown in Table~\ref{tab:class_stats}.

\subsection{Model Architecture and Training}

All computation was performed using NVIDIA GeForce RTX 2060. We trained the model over 100 epochs using a batch size of 32, an initial learning rate of 0.001, AdamW optimisation~\cite{Loshchilov2017-bk}, and cross-entropy loss $-\sum_{c=1}^{C}\tau_{c}\log{(\rho_{c})}$ on the $\text{Softmax}(x_{i}) = \frac{e^{x_{i}}}{\sum_{c=0}^{C} e^{x_{c}}}$ output. This protocol was designed after significant experimentation, considering a range of architectures and hyperparameters. We found that deeper encoders and other segmentation models provided in the Segmentation Models Pytorch library~\cite{Yakubovskiy2019} provided either no significant increase proportional to computation cost, or a decrease in model performance. We also explored the use of focal loss~\cite{Lin2020-po} to address class imbalance, but found no improvement in performance.

\section{Results}

To evaluate model performance for each class we used a number of established metrics for semantic segmentation tasks of this type as shown, comprising precision $P$, recall $R$ and F1 score $F$, where target $\tau \in \{0,1\}$ is the $d \times d$ pixel-wise ground truth map for that class, and output $\rho \in [0,1]$ the $d \times d$ softmaxed model prediction.

These metrics were also applied using only the centroid coordinates of each cell -- due to the nature of the task at hand, identifying which proteins are present and their location is more relevant than perfectly segmenting each individual cell, and as such we do not wish to unduly penalise the model if it correctly locates and classifies a protein expressing cell, but fails to capture the exact shape of that cell.

\begin{equation}
    \begin{aligned}
        &tp = \sum_{x,y=0}^{d} \tau_{x,y}\rho_{x,y} \\
        &tn = \sum_{x,y=0}^{d} (1 - \tau_{x,y})(1 - \rho_{x,y}) \\
        &fp = \sum_{x,y=0}^{d} \rho_{x,y}(1 - \tau_{x,y}) \\ 
        &fn = \sum_{x,y=0}^{d} \tau_{x,y}(1 - \rho_{x,y}) \\
        &P = \frac{tp}{tp+fp} \qquad R = \frac{tp}{tp+fn} \qquad F1 = \frac{2PR}{P + R} \\
    \end{aligned}
\end{equation}

\begin{table}
    \centering
    \begin{tabular}{|c|c|c|c|}
        \hline
        Per-Pixel & F1 & Precision & Recall \\
        \hline
        CD8\_CD3LO & 0.96 / 0.92 & 0.96 / 0.95 & 0.96 / 0.90 \\
        CD3 & 0.96 / 0.91 & 0.96 / 0.90 & 0.96 / 0.92 \\
        CD20 & 0.97 / 0.93 & 0.96 / 0.96 & 0.97 / 0.90 \\
        CD8\_CD3HI & 0.95 / 0.89 & 0.95 / 0.88 & 0.95 / 0.91 \\
        Avg. & 0.96 / 0.91 & 0.96 / 0.91 & 0.96 / 0.91\\
        \hline
        \hline
        Per-Centroid & F1 & Precision & Recall \\
        \hline
        CD8\_CD3LO & 0.99 / 0.96 & 0.99 / 0.98 & 0.99 / 0.94 \\
        CD3 & 0.99 / 0.96 & 0.99 / 0.95 & 0.99 / 0.96 \\
        CD20 & 1.00 / 0.97 & 1.00 / 0.99 & 1.00 / 0.95 \\
        CD8\_CD3HI & 0.99 / 0.92 & 0.99 / 0.91 & 0.99 / 0.95 \\
        Avg. & 0.99 / 0.95 & 0.99 / 0.96 & 0.99 / 0.95 \\
        \hline
    \end{tabular}
    \caption{Mean per-pixel and per-centroid segmentation performance on training set / holdout test set.}
    \label{tab:metrics}
\end{table}

\begin{figure}[h!]
\centering
\includegraphics[width=\textwidth]{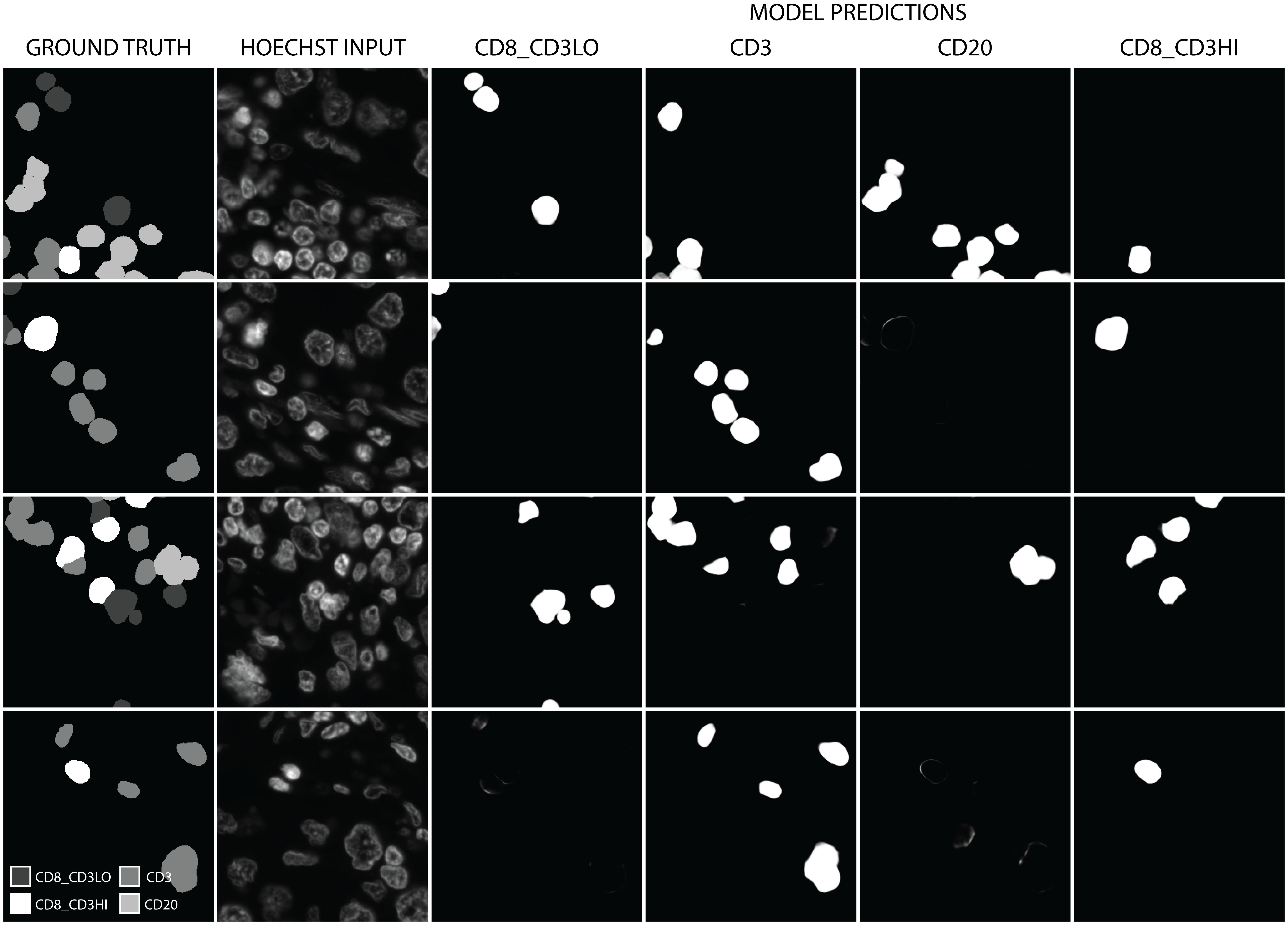}
\caption{Examples of model predictions on test set images -- here we show the ground truth segmentation mask of class labels, the Hoechst 33342 stained input image, and the predicted segmentation output for each class.}
\label{fig:preds}
\end{figure}

Table~\ref{tab:metrics} shows the performance of our model according to these metrics. We achieve over 90\% precision, recall and F1 score across all classes on the test set, showing excellent generalisation ability to unseen data with only a small decrease in performance compared to the training set. CD8\_CD3HI expressing cells were the most difficult to classify, likely due to their lower representation across the training set. As expected, we see a small increase in performance across these metrics when evaluating the cell centroid classifications only, because in this case imperfect segmentation of regions at the edges of cells is disregarded. Figure~\ref{fig:preds} shows four random sample pairs from the test set, along with the model output predictions for each class -- as we can see here, the segmentation is reliably good.

\section{Discussion}

\begin{figure}
\centering
\includegraphics[width=\textwidth]{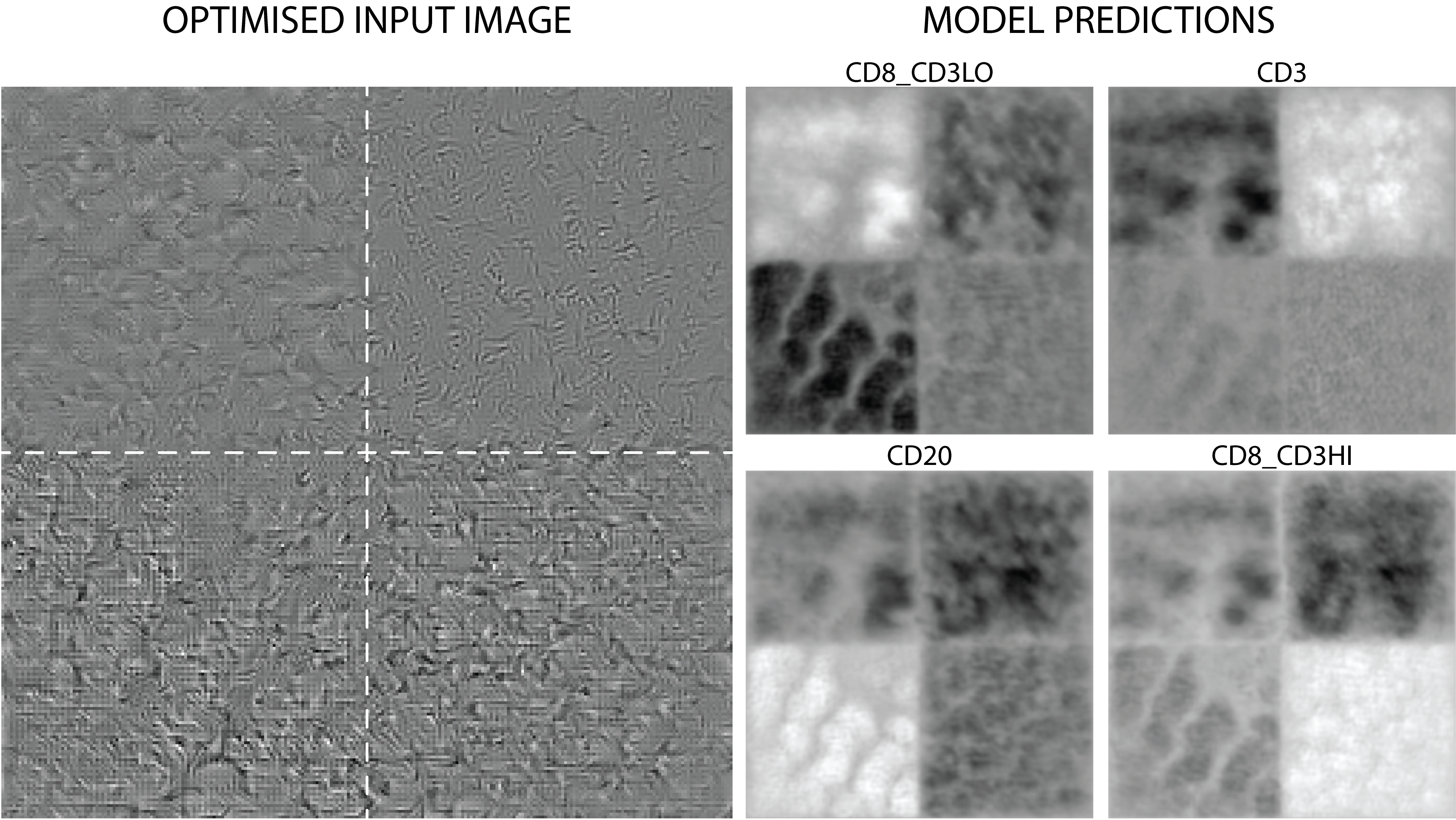}
\caption{Optimised input image generated using stochastic gradient descent to maximise class predictions for each quadrant.}
\label{fig:img_opt}
\end{figure}

To explore how the network is able to distinguish between different lymphocytes so successfully, we use stochastic gradient descent (SGD) to optimize an input image $I$ set uniformly to the normalised mean of all training set images to maximise output logits (raw outputs, without activation or normalisation) for each class, as proposed by Simonyan et al.~\cite{Simonyan2013-rl}. We extend this technique for multi-class segmentation using a custom loss function designed to balance per-class optimisation for class-specific quadrants in a single image, allowing for easy comparison of learned features. 

\subsection{Feature Visualisation}
For input image optimisation we used a learning rate of 1.0 and ran SGD for 10,000 steps. We optimise each quadrant for a different positive class (i.e. those classes not consisting of background or cells outwith our classes of focus), by specifying a mask $\mu$ of dimension $C \times d \times d$ (where $d$ is the $x$ and $y$ dimension of the input image -- in this case 256 pixels -- and $C$ is the number of positive classes -- in this case 4) such that for each class $c~\text{in}~C$ a different quadrant is set to ones, and the remaining elements in $c$ to zero -- the objective being to maximise the model outputs for each class in a specific region, as shown in Figure \ref{fig:img_opt}. This mask is used to separate pixels to maximise and those to minimise in the model outputs $\rho$ according to class. The Rectified Linear Unit (ReLU) is applied to the regions to minimise, to control for disproportionate minimisation of negative regions in preference to the positive regions, in which we are interested. We also use a regularisation term consisting of the variance between positive class outputs to ensure that the input is not disproportionately maximised for classes associated with simpler features, as shown below. Simonyan et al.\ included an additional $L_{2}$ regularisation term for the input image $I$ to counteract over-optimisation for a single feature, but in practice we found that this lead to negligible change in output and so we omit it -- likely this omission is possible because we are optimising for mean pixel classifications and thereby for multiple features, rather than for a single label.

\begin{equation}
    \begin{aligned}[left]
    &ReLU\left(x\right)=~
     \begin{cases}
      & 0~\text{if}~x < 0 \\ 
      & x~\text{otherwise} \\
    \end{cases}
    \\
    &V\left (x \right ) =~ \frac{\sum_{i}^{}\left ( x_{i} - \bar{x}\right )^{2}}{C}\\
    &Loss =~\frac{1}{d^2C}\sum_{c=0}^{C}\sum_{x,y=0}^{d} ReLU \left ( \rho_{c,x,y} \left | \mu_{c,x,y}-1 \right | \right ) - \mu_{c,x,y} \rho_{c,x,y} \\
    &~~~~~~~+ V \left ( \left [\frac{1}{d^2}\sum_{x,y=0}^{d} ReLU \left ( \rho_{c,x,y} \left | \mu_{c,x,y} - 1 \right | \right ) - \mu_{c,x,y} \rho_{c,x,y} ~ for ~ c = 1\dots C \right ] \right ) \\
    \end{aligned}
\end{equation}

As shown in Figure~\ref{fig:img_opt}, the results of this experiment show that the model has learned textural features of the nuclear chromatin to discriminate between different classes, as the predominant difference between regions is textural and evident features are small. This difference in texture is particularly pronounced between CD8\_CD3LO and CD3 only expressing cells, with the CD3 only class quadrant in particular optimising for quite small, curved features. In models like ours where the output class probabilities for each pixel are calculated -- and thereby constrained and exaggerated -- through use of softmax as an output function, it is difficult to differentiate a strong positive signal for an individual class from a strong negative signal for the \emph{other} classes. This also holds for the logits which we here hope to maximise for each class -- in the context of actual predictions by the model, these logits would be passed into a softmax function, and as such interact in a way we so not visualise here -- for example, classes A and B may have a high logit output, but if A is lower than B, the softmaxed output will strongly favour B even if the raw difference in output was not great. This makes optimisation-based visualisation of features in this way problematic: if we optimise for softmaxed outputs, the output is quickly saturated and the inputs are unintelligible, but if we optimise for logits, we do not accurately represent the actual classifications of the entire model.

\begin{figure}[h!]
\centering
\includegraphics[width=\textwidth]{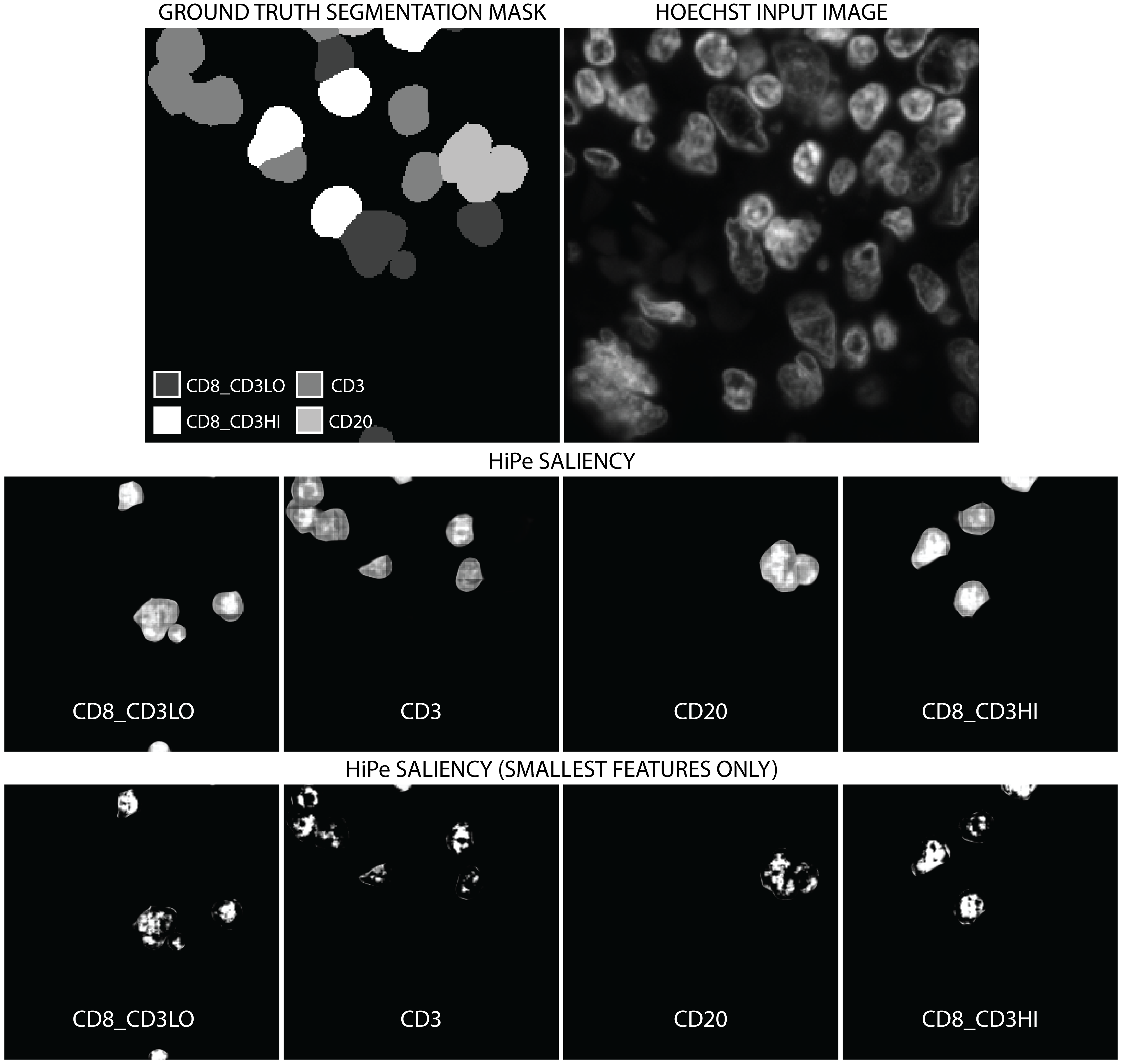}
\caption{Labelled immune cell mask from multiplex immunofluorescence image (top left) and corresponding Hoechst 33342 stained patch (top right). Below, HiPe~\cite{Cooper2021-km} saliency maps for each.}
\label{fig:hipe}
\end{figure}

To gain further insight into the learned features, we used Hierarchical Perturbation~\cite{Cooper2021-km} (HiPe) to generate saliency maps -- visualisations of which regions of the input image were more or less important in determining the ultimate output of the model -- at different resolutions, as shown in Figure~\ref{fig:hipe}. We used the standard implementation of HiPe on the softmaxed output of the model, with ``fade'' perturbation (such that occluded portions are replaced with zero input) and four initial cells. Input saliency based methods like HiPe are more transparently interpretable than input optimisation, as they explicitly show which areas of the input image were more or less important in determining the output for each class. We use HiPe in preference to other input saliency based explanatory techniques as it is much quicker than similar perturbation-based saliency methods for multiple large images containing relatively small salient features, as is the case with WSIs, and is more precise than gradient-based methods which are often indistinct. 

We found that larger salient regions were comprised of the cells themselves, as would be expected -- but more interestingly, that the most salient regions were smaller, appearing to cluster in the nuclei of the salient cells. This supports our conclusion that the model is using morphological features of the chromatin made visible by Hoechst 33342 staining to perform the classification. We also note that the model does not find regions outwith the cells in question salient, showing that the proximity or morphology of nearby cells or tissue structures is not used to inform the segmentation at all.

In this work we show that with new explicability techniques, neural networks can be valuable tools for discovery as well as for automation. We demonstrate for the first time that it is possible to identify different lymphocyte protein expressions from morphological features only and present a deep learning methodology to achieve this, which we hope will have significant impact in digital pathology.

We intend to apply proven prognostic metrics such as immune cell infiltration evaluation to slides labelled using our method, with the goal of drastically reducing the cost of immune profiling and thereby allowing more patients to benefit. Future work will include exploring semi-supervised and unsupervised approaches to classification via clustering to reduce labelling burden when training new models, alongside extending and applying our approach to other cancers and proteins.

\subsection{Code Availability}
All code is available at github.com/jessicamarycooper/ICAIRD.

\subsection{Data Availability}
The data that support the findings of this study are available from the corresponding author upon reasonable request.

\section{Acknowledgements}
This work is supported by the Industrial Centre for AI Research in digital Diagnostics (iCAIRD) which is funded by Innovate UK on behalf of UK Research and Innovation (UKRI) [project number: 104690] and NHS Lothian.

\section{Author Contributions}
All authors contributed to conceptualisation, analysis, reviewing and editing. OA and DH were responsible for funding acquisition and resources, administration and supervision. Data curation, imaging methodology and investigation were done by JC and IU. IU wrote the Multiplexed Immunofluorescence (mIF) Protocol. JC wrote the software and the remainder of the paper. Deep learning methodology and visualisation was done by JC.

\section{Competing Interests}
The authors declare that there are no competing interests.

\bibliographystyle{plain}
\bibliography{references}
\end{document}